\documentclass{article}

\usepackage{microtype}
\usepackage{graphicx}
\usepackage{subfigure}
\usepackage{booktabs} 

\usepackage{hyperref}

\usepackage[accepted]{icml2025}

\usepackage{amsmath}
\usepackage{amssymb}
\usepackage{mathtools}
\usepackage{amsthm}

\usepackage[capitalize,noabbrev]{cleveref}

\theoremstyle{plain}

\theoremstyle{definition}

\theoremstyle{remark}

\usepackage[textsize=tiny]{todonotes}

\icmltitlerunning{Sparsity Meets Similarity}

\begin{document}

\twocolumn[
\icmltitle{Sparsity Meets Similarity: Leveraging Long-Tail Distribution for Dynamic Optimized Token Representation in Multimodal Large Language Models}



\icmlsetsymbol{equal}{*}

\begin{icmlauthorlist}
\icmlauthor{Gaotong Yu}{equal}
\icmlauthor{Yi Chen}{equal,yyy,comp}
\icmlauthor{Jian Xu}{yyy,comp}
\end{icmlauthorlist}

\icmlaffiliation{yyy}{School of Artificial Intelligence, University of Chinese Academy of Sciences, Beijing 100049, China}
\icmlaffiliation{comp}{State Key Laboratory of Multimodal Artificial Intelligence Systems (MAIS), Institution of Automation, Chinese Academy of Sciences, Beijing 100190, China}

\icmlcorrespondingauthor{Gaotong Yu}{yogurt2077@gmail.com}
\icmlcorrespondingauthor{Yi Chen}{yi.chen@nlpr.ia.ac.cn}
\icmlcorrespondingauthor{Jian Xu}{jian.xu@ia.ac.cn}

\icmlkeywords{Machine Learning, ICML}

\vskip 0.3in
]



\printAffiliationsAndNotice{\icmlEqualContribution} 

\begin{abstract}
 Recently, multimodal large language models (MM-LLMs) have achieved significant success in various tasks, but their high computational costs limit widespread application. The main computational burden arises from processing concatenated text and visual tokens in the LLM layer, where input token length directly affects efficiency. Our analysis of visual tokens reveals that their similarity to the CLS token follows a long-tail distribution, with only a few showing high similarity. To address this, we propose a dynamic pruning algorithm that identifies the inflection point in the visual CLS token similarity curve, enabling effective trimming of visual markers to accelerate model performance. Additionally, we perform a second round of pruning in the LLM layer, filtering out low-correlation tokens through the interaction between visual and textual features. Experimental results demonstrate that our method achieves performance comparable to the original while utilizing only 22\% of the original token quantity. Our source code will be made publicly available upon acceptance.

\end{abstract}

\section{Introduction}
Benefiting from the training on massive text data and the continuous development of deep neural network structures, as well as the methods of pre-training and fine-tuning, large language models (LLMs) \cite{brown2020language, ouyang2022training, jiang2023mistral, touvron2023llama} have become the forefront of research in the field of artificial intelligence. Despite the remarkable achievements of deep learning methods in various tasks in recent years, the emergence of LLMs has sparked a revolutionary breakthrough in the field of artificial intelligence.
\begin{figure}[ht]
\centering
\includegraphics[width=1.\linewidth]{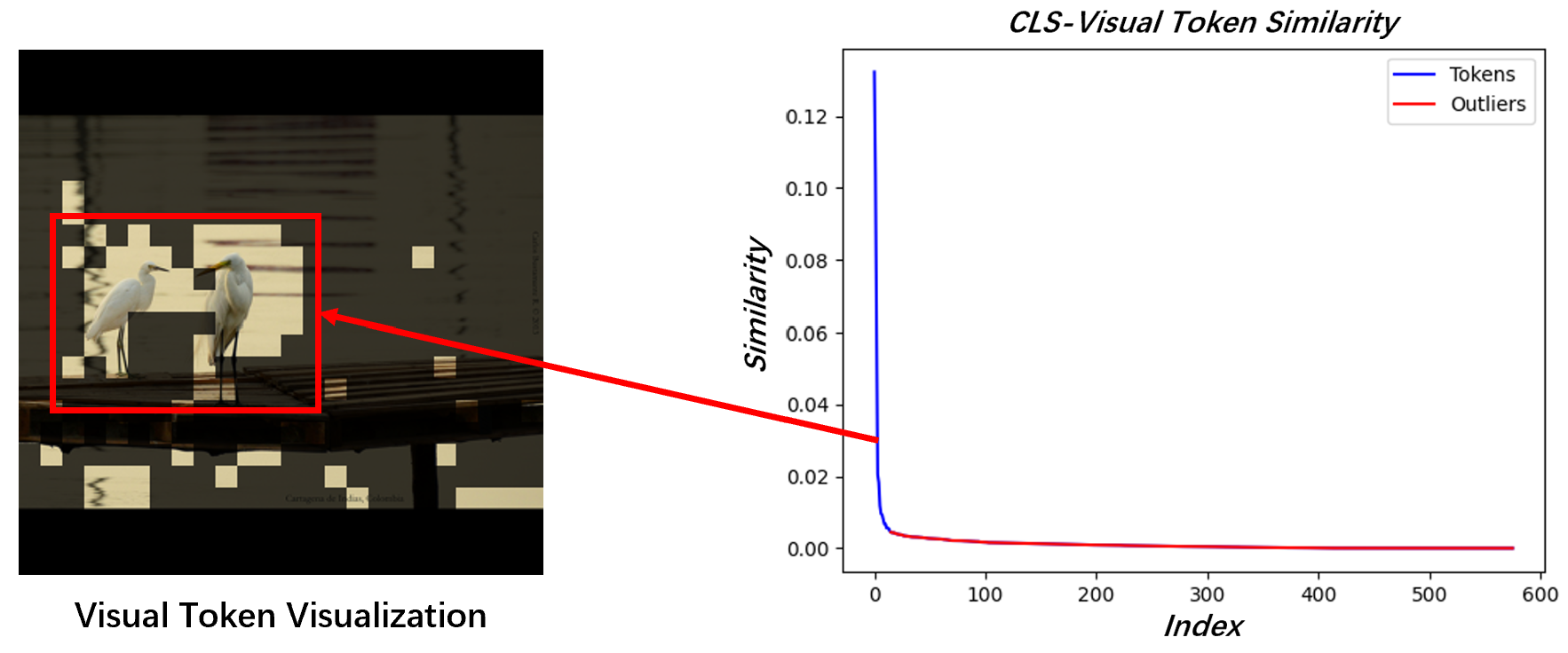}
\caption{Visual tokens have sparsity. Visualize CLS visual token similarity, with black areas representing the `tails', and others representing the `heads". The red boxes represent key areas of the entire image.} 
\label{fig: long}
\vspace{-3mm}
\end{figure}  

Utilizing large language models as the core component, multimodal large language models (MM-LLMs) \cite{liu2023improved, liu2024llava, openai2023gpt, team2023gemini} take both visual and textual information as input to address multimodal tasks. Specifically, MM-LLMs employ a visual encoder, such as Vision Transformer (ViT) \cite{dosovitskiy2020image} and CLIP \cite{radford2021learning}, and mapping layers, such as Multi-Layer Perceptions (MLP), to extract visual features and transform them into textual feature space. They then leverage the powerful language generation, zero-shot learning, and contextual learning capabilities of LLMs to process concatenated visual and textual features. MM-LLMs demonstrate performance surpassing traditional models across various multimodal tasks, showcasing remarkable emergent capabilities.

Due to the enormous number of parameters in MM-LLMs, both training and inference require substantial computational resources, with LLMs predominantly driving the computational costs due to their larger parameter count compared to the visual encoder. Furthermore, since LLM mostly employs transformer architectures, their computational cost often quadratically increases with input length. Therefore, the efficiency of MM-LLMs is significantly influenced by the number of visual tokens generated by the visual encoder. In addressing the issue of enhancing LLM efficiency, two widely used programmes are currently employed. The first program involves replacing high-parameter LLM with smaller-parameter LLM. By directly reducing the model parameters, significant reductions in computational requirements can be achieved, thereby speeding up both training and inference. For instance, Chu et al \cite{chu2023mobilevlm}. proposed a mobile device MM-LLM by replacing the LLM with $1.4B$ and $2.7B$ parameters model. To achieve local deployment, Yuan et al \cite{yuan2023tinygpt}. proposed the TinyGPT-V, based on small backbones with a $2.8B$ parameter. Although replacing the original large parameter LLM with lightweight LLM can improve the computational efficiency and local deployment of the model, this method will reduce the inference ability of the model, resulting in a significant decrease in performance.

The second program involves reducing the output of the visual encoder to decrease the input length of LLM, effectively enhancing the overall training and inference speeds of the model. Recently, there have been related works that use this to achieve efficient inference of MM-LLMs. For example, Wang et al \cite{wang2023smarttrim}. designed a lightweight module for identifying and removing redundant tokens and attention heads in each layer. By adding this module to the original model, the training and inference process of the model can be accelerated. Shi et al \cite{shi2023crossget}. presents Cross-Guided Ensemble of Tokens (CrossGET), a versatile acceleration framework designed for vision-language Transformers. By dynamically merging tokens during inference, CrossGET effectively lowers computational expenses without compromising performance quality.
Shang et al \cite{shang2024llava}. proposed an adaptive token pruning strategy based on the LLaVA framework, which reduces the output of visual encoders by clustering visual tokens, thereby improving the computational efficiency of LLaVA. Building upon this, Chen et al \cite{chen2024recoverable}. designed a visually guided token recovery mechanism based on textual information, which recovers important information that may be discarded by calculating the similarity between the question text and the visual tokens.

Inspired by the second method, we have designed a dynamic optimization algorithm for token compression targeting long-tail distribution. Specifically, we explored the distribution of visual tokens. As shown in Figure \ref{fig: long}, when arranging the similarity between the CLS and the visual token in descending order, we observed a long-tail distribution in the similarity curve. This implies that there is a significant amount of redundancy in visual tokens, with only a small portion of visual tokens highly correlated with the input image.
Therefore, we have devised an objective function that compresses visual tokens by aiming to find a segmentation point between the ``head" and ``tail" of the long-tail distribution. By optimizing this objective function, we can reduce the output of the visual encoder. Subsequently, to further accelerate the model, we have adopted a method similar to Zhang et al \cite{zhang2024h2o}, where pruning is performed by calculating the importance between mapped visual tokens and textual tokens. During this process, there is an interaction between visual and textual information, and the pruning process can be seen as extracting tokens with high visual-textual relevance. This ultimately leads to a further reduction in the input sequence length of the LLM, enhancing the inference efficiency of MM-LLMs. At the same time, due to the different distribution of visual tokens for different tasks and instances, we have added a dynamic pruning rate setting to dynamically process each sample/task during the pruning process, ensuring that the model can achieve a balance between efficiency and performance on different data.

Our main contributions are summarized as follows:
\begin{itemize}
    \item We have developed a dynamic optimization algorithm for token compression that targets long-tail distributions. With this approach, we achieve competitive performance, delivering a compression ratio of up to 8x on most datasets.
    \item Based on the different input instances, we have proposed the dynamic pruning rate to ensure dynamic pruning for each sample/task, thus ensuring the generalization of the algorithm.
    \item Our method is simple and efficient. It can be directly integrated into the model for training-free inference, and it can also participate in training alongside the model during the fine-tuning phase. Meanwhile, we have conducted extensive experiments on various multimodal models to validate the effectiveness and compatibility of the method.
\end{itemize}

\section{Related Work}
\subsection{Visual Token Pruning Method}
In the MM-LLMs framework, visual encoders commonly leverage the ViT framework. Presently, numerous researchers are dedicated to enhancing the efficiency of ViT models, with a focus on token compression emerging as a key area of study. Token compression strategies encompass both token pruning and token merging techniques. 

The former evaluates the importance of different tokens based on defined criteria, retaining significant tokens while discarding insignificant ones. For instance, Rao et al \cite{rao2021dynamicvit}. introduced a dynamic pruning technique that gradually and adaptively prunes redundant tokens at each model layer by considering the visual attention sparsity. This method estimates the importance scores of individual tokens using current features. Going beyond this, Kong \cite{kong2022spvit} and Xu \cite{xu2023no} proposed that insignificant tokens should not be merely discarded but instead incorporated or subject to additional modifications to prevent the permanent loss of image data resulting from improper pruning. Wu et al. \cite{wu2023ppt} propose the token Pruning \& Pooling Transformers (PPT), which heuristically integrates both token pruning and token pooling techniques in ViTs without additional trainable parameters. Endo et al \cite{endo2024feather} propose FEATHER (Fast and Effective Acceleration wiTH Ensemble cRiteria), a straightforward approach, Which has more than 5× performance improvement on the vision-centric localization benchmarks compared to the original acceleration approach.

The latter involves grouping similar tokens together, discarding unimportant background tokens, and achieving efficient token compression by merging foreground tokens. Bolya et al. \cite{bolyatoken} introduced token merging (ToMe), a simple method to increase the throughput of existing ViT models without training. Chen et al \cite{chen2023diffrate}. linked the loss function with the compression rate to autonomously determine varied token compression rates across different layers, employing a combination of pruning and merging techniques simultaneously. Long et al \cite{long2023beyond}. ensured the merging process's reliability by factoring in token importance and diversity for pruning, subsequently merging akin tokens. Similarly, Lee et al \cite{lee2024multi}. underscored the significance of considering diverse token relationships during merging processes. Han et al \cite{han2024rethinking} propose a unified ``filter-correlate-compress'' paradigm that decomposes the token reduction into three distinct stages within a pipeline, maintaining consistent design objectives and elements while allowing for unique implementations. 

\subsection{Efficient Large Language Model}
Large language models have garnered significant attention due to their outstanding performance across various tasks. However, the substantial computational and memory requirements for inference with large language models pose challenges for their deployment in resource-constrained environments. Researchers have been diligently working on developing technologies aimed at enhancing the efficiency of inference with large models.

Methods to enhance the efficiency of large models can be broadly categorized into three levels of optimizations: data-level, model-level, and system-level \cite{zhou2024survey}.
Data-level optimization improves efficiency by optimizing input prompts, without altering the original model, thereby avoiding high model training costs. For example, Jiang et al \cite{jiang2023llmlingua}. divided the prompt into three parts, and then used additional models to calculate the perplexity of each part separately. Based on the perplexity, prompts were removed to accelerate the model. On this basis, Jiang et al \cite{jiang2023longllmlingua}. proposed using question text to compress the document section, further improving the computational efficiency of the model.

Model-level optimization is typically carried out during the model inference process, by designing effective model structures or compressing pretrained models to improve efficiency. Usually, it requires pretraining or fine-tuning to preserve or restore model functionality. Such as Zhang et al \cite{zhang2024h2o}. found that only a small number of tokens made outstanding contributions during the attention process. Based on this, they designed a dynamic KV cache eviction policy that balances recent and important tokens. Li et al \cite{li2024snapkv}. discover that each attention head in the model consistently focuses on specific prompt attention features during generation. Drawing on this insight, they proposed the SnapKV, which automatically compresses KV caches by selecting clustered important KV positions for each attention head.

System-level optimization involves optimizing the inference engine or service system. Optimization of the inference engine does not require model training, and optimization of the service system is non-destructive to model performance.

Building upon the works above, we have designed a dynamic optimization algorithm for token compression targeting long-tail distribution.
Initially, we observed a long-tail distribution of similarity between the visual and the CLS token generated by the visual encoder. This suggests that not all tokens contain valuable information for the input image. Hence, we developed a dynamic segmentation function to locate the split point between the ``head" and ``tail" of this long-tail distribution, extracting important tokens in the visual modality for pruning, ensuring efficiency, and maintaining high performance. Subsequently, at the LLM layer, we computed the importance of visual tokens aligned with the text modality and text tokens mutually. Based on this evaluation, we pruned the inputs to the LLM layer once more, striking a balance between efficiency and performance. Our approach dynamically processes each input instance, showcasing strong generalization capabilities.

\section{Method}
\subsection{Overview}
In order to improve computational efficiency while ensuring performance availability, we have designed a dynamic optimization algorithm for token compression targeting long-tail distribution as shown in Figure \ref{fig: overview}. This method consists of the following three steps:

\begin{figure*}[!htb]
\centering
\includegraphics[width=0.85\linewidth]{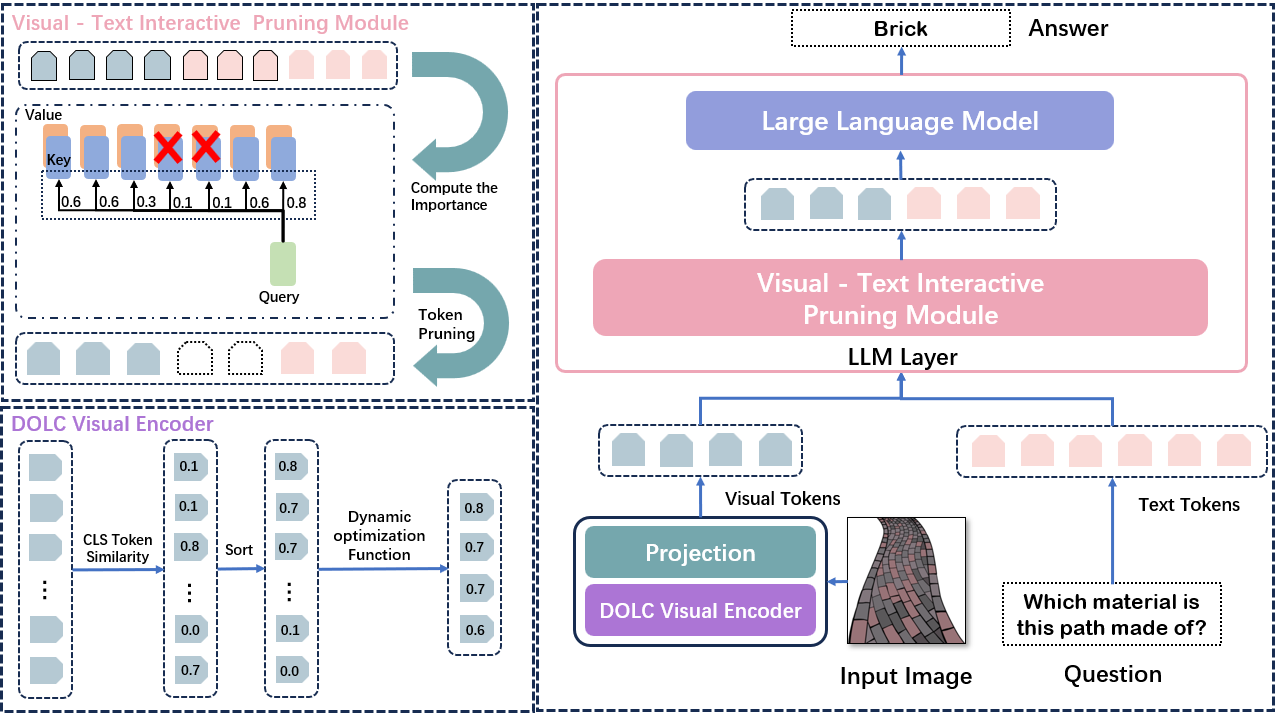}
\caption{The framework of dynamic optimization algorithm for token compression targeting long-tail distribution, where the right side is the overall framework and the left side is the submodule framework diagram.} 
\label{fig: overview}
\end{figure*}  

\begin{itemize}
    \item \textbf{Dynamic Segmentation Targeting Long-tail Distribution\quad} Firstly, calculate the similarity between CLS and visual tokens and sort the visual tokens according to their similarity. Then, maximize the dynamic optimization function targeting long-tail distribution to extract tokens with high CLS-visual similarity and discard other tokens to complete the pruning of the visual tokens.
    \item \textbf{Token mapping alignment and concatenation\quad} The pruned visual token is aligned to the feature space of the textual token through a mapping layer. And concatenate them with the text tokens.
    \item \textbf{Token pruning based on visual-text interaction\quad} Compute the similarity between each token and its preceding token, using this as the importance score for each token. Sort based on this importance and remove the lowest-scoring N tokens to reduce the input length at the LLM layer. During the calculation of importance scores, for each current token, the similarity with all preceding tokens is considered. Therefore, for every text token, similarity with all visual tokens is computed, biasing towards retaining tokens with high visual-text correlation.
\end{itemize}

After completing the above steps, send the token to LLM to obtain the final output.

\subsection{Dynamic Segmentation Algorithm Targeting Long-tail Distribution}
\begin{figure}[!htb]
\centering
\includegraphics[width=1.\linewidth]{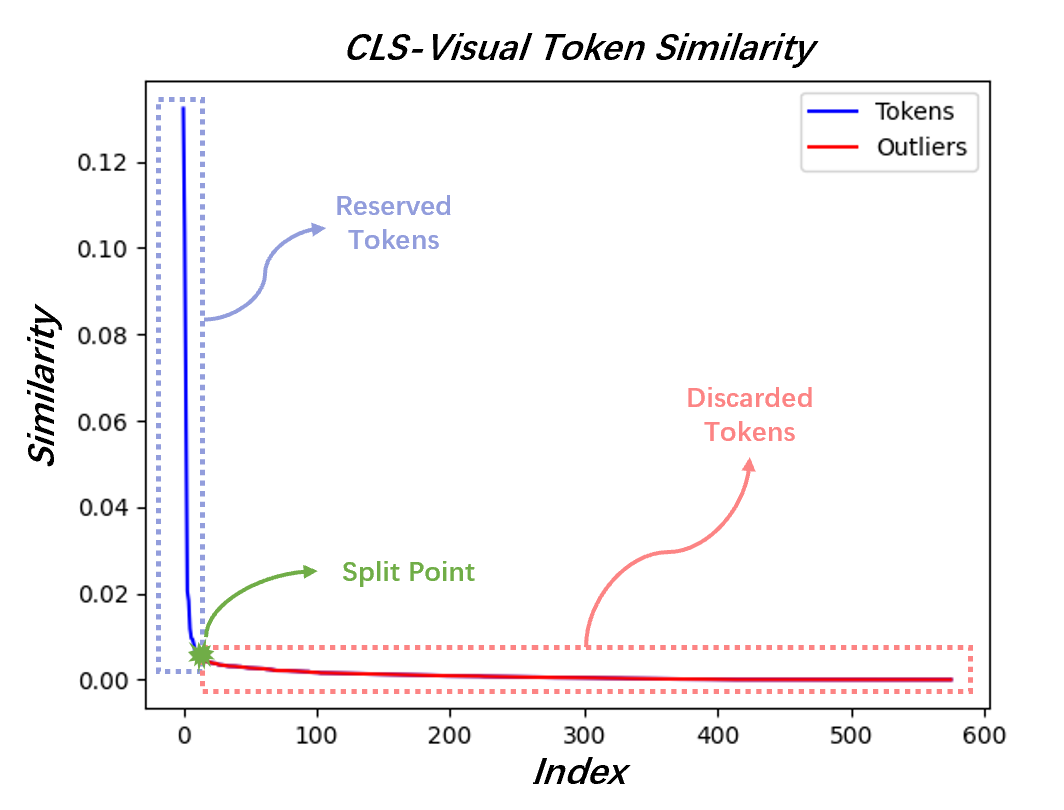}
\caption{Example of converting token pruning task into long tail distribution segmentation task.} 
\label{fig: c1}
\end{figure}

Inspired by the works of Zhang \cite{zhang2024h2o} and Li \cite{li2024snapkv}, we conducted a study on the similarity between the CLS and visual tokens generated by the visual encoder. Our research revealed the sparsity of similarity between the CLS and visual tokens, indicating that only a few tokens exhibit high similarity. Therefore, we arranged the similarity scores of CLS-visual token pairs in descending order. As shown in Figure \ref{fig: c1}, the curve demonstrates a long-tail distribution, suggesting that it is possible to retain only the ``head" portion containing tokens with high similarity, while discarding the ``tail" portion with low similarity to reduce the output length of the visual encoder and improve overall computational efficiency.

To segment the long-tail distribution curve effectively, firstly, calculate the similarity between CLS and visual tokens, and the similarity formula is as follows: 
\begin{equation}
    Similarity = Softmax(\frac{_{cls \ token} \cdot W_{visual \ tokens}^T}{\sqrt{d_{W_{cls \ token}}}}).
\end{equation}

Then, arrange the similarity in descending order, the CLS-visual similarity set can be obtained as$\mathrm{D}=\left\{\mathrm{d}_1, \mathrm{~d}_2, \ldots,\mathrm{d}_{\mathrm{n}}\right\} \text {, wherein} \textbf{ } \mathrm{d}_1 \geq \mathrm{d}_2 \geq \ldots \geq \mathrm{d}_{\mathrm{n}}$ and $N$ is the number of tokens. 

The definition of the splitting objective function is as follows:
\begin{equation}
f(i)=\frac{(n-i) \cdot\left(d_1-d_i\right)}{c}.
\label{qiefen}
\end{equation}
wherein $c=d_1 - d_n$ and $i \in [1,n-1]$.

To obtain the optimal segmentation point (i.e. the segmentation position of the "head" and "tail"), it is necessary to solve for the solution $i^*$ that maximizes the Equation \ref{qiefen}.
The above can be defined as:
\begin{equation}
i^*=\operatorname{argmax}_{i \in\{1,2, \ldots, n-1\}}\left(\frac{(n-i) \cdot\left(d_1-d_i\right)}{d_1-d_n}\right).
\label{arg}
\end{equation}

The meaning of this Equation \ref{qiefen} and \ref{arg} is as follows: 
\begin{itemize}
    \item For each possible segmentation point $i$ (from $1$ to $n-1$), we calculate $f(i)$. 
    \item The molecule $(n-i) \cdot (d_1 - d_i)$ of $f(i)$ represents $(n-i)$ is the number of data points to the right of the segmentation point (including the segmentation point) and $(d_1- d_i)$ is the difference between the maximum value and the segmentation point value.
    \item The denominator $(d_1 - d_n)$ is a constant used for normalization.
    \item We search for the value of $i$ that maximizes this ratio, which is the optimal segmentation point.
\end{itemize}

\begin{figure}[!htb]
\centering
\includegraphics[width=0.85\linewidth]{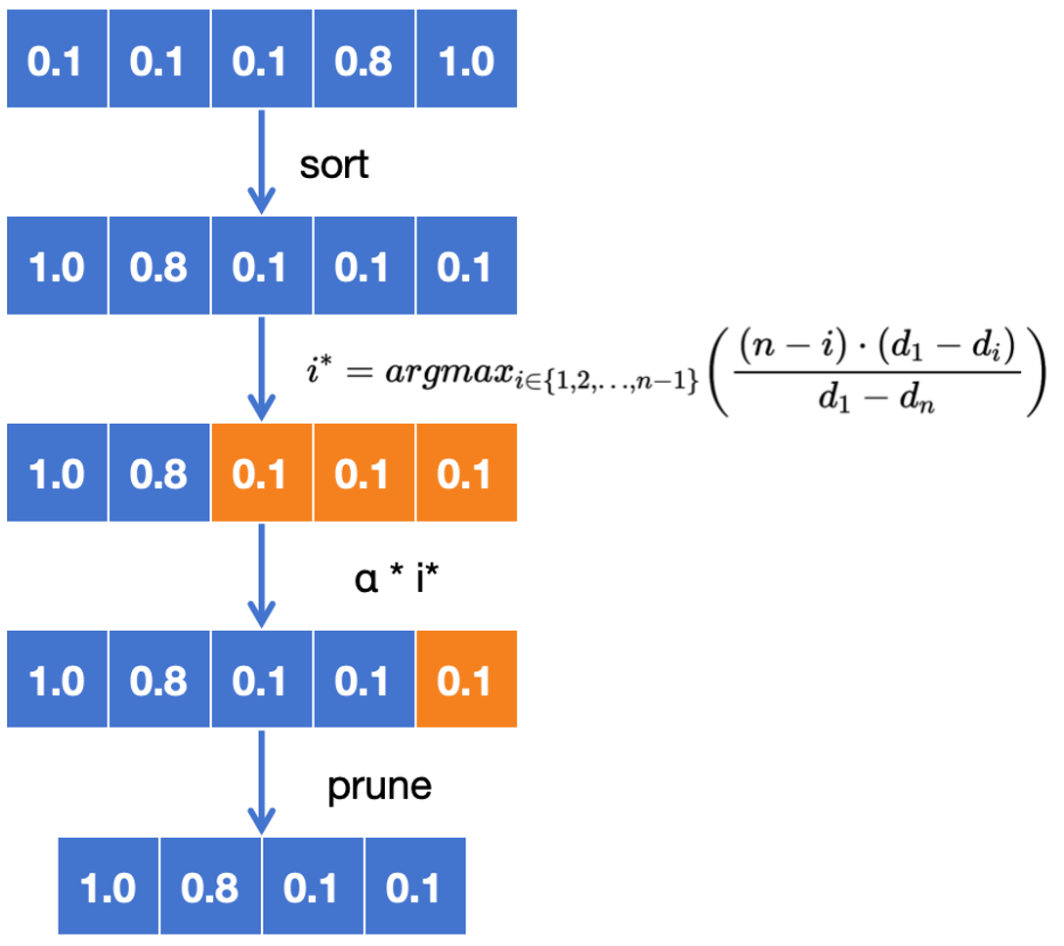}
\caption{Example of visual token pruning with our method.} 
\label{fig: step}
\end{figure}  

To prevent gradient collapse during model training, $i^*$ is multiplied by a smoothing coefficient $\alpha$ in practical implementation. The overall steps are shown in  Figure \ref{fig: step}. After the operation above, the pruning of the visual modality is completed. It is important to note that this process is dynamic for each input instance. This means that for different inputs, we will individually solve the objective function based on the distribution of their visual tokens. This dynamic approach for each sample balances efficiency and effectiveness by providing targeted processing for each instance.

\subsection{Visual-Text Interactive Token Pruning Algorithm}
After pruning the output of the visual encoder, inspired by \cite{zhang2024h2o}, we adopt a similar approach to further enhance the computational efficiency. We will perform a second round of token pruning before feeding into LLM.

\begin{figure}[!htb]
\centering
\includegraphics[width=1.\linewidth]{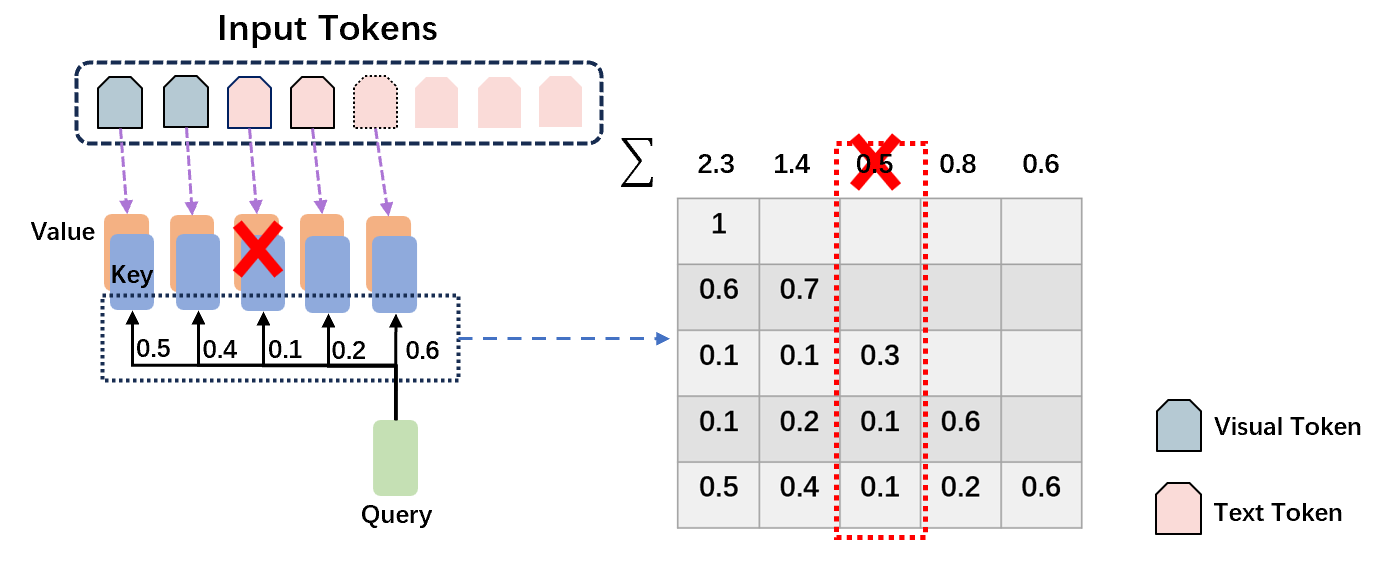}
\caption{Example of visual-text interactive pruning.} 
\label{fig: text}
\end{figure}  

Firstly, the pruned output of the visual encoder will be passed through a projection layer to map it to the feature space of text tokens. Next, as shown in Figure \ref{fig: text}, the mapped visual tokens will be concatenated with text tokens. Subsequently, the similarity between each token and all preceding tokens will be calculated, where the importance score of each token is the sum of its similarity scores with other tokens. Finally, based on the importance scores, tokens with low importance scores will be removed. 
The specific drop steps are as follows:
\begin{itemize}
    \item Set the recent tokens $recent \ budget$ that need to be saved and the remaining tokens $heavy \ budget$ that need to be retained, where the number of $recent \ budget$ is $M$ and the number of $heavy \ budget$ is $N$.
    \item Keep the last $M$ tokens without dropping.
    \item When the current number of tokens $X$ does not exceed $M+N$, continuously accumulate the score of $heavy \ budget$.
    \item When the current number of tokens $X>M+N$, start dropping $heavy \ budget$.
    \item Continue to accumulate $heavy \ budget$. Once the number of tokens $X>M+N$, continue the above process.
\end{itemize}

During this process, the similarity between visual tokens will be calculated once again to complete the second round of pruning in the visual modality. Additionally, each text token will undergo similarity calculations with preceding visual tokens. This implies that visual tokens and text tokens will interactively compute each other's importance during this process, guiding the model to retain tokens with high similarity between visual and text modalities, thereby further enhancing the computational efficiency of the model.

\section{Experiments}
\subsection{Dataset}
We assessed the effectiveness of our approach on several well-known and publicly available multimodal datasets, including \textbf{ScienceQA} \cite{lu2022learn}, \textbf{TextVQA} \cite{singh2019towards}, \textbf{MME} \cite{fu2023mme}, \textbf{VQAv2} \cite{goyal2017making}, \textbf{POPE} \cite{li2023evaluating}, and \textbf{MMBench} \cite{liu2023mmbench}.

\subsection{Implementation Details}
All experiments were conducted in the PyTorch framework on four NVIDIA 4090 24G GPUs. We utilized the \textbf{InterVL2-1B}, \textbf{Phi3v}, \textbf{LLaVA1.5-7B} and \textbf{LLaVA1.6-Next} as our baseline. Set hyperparameter $\alpha$ to 0.24 in dynamic visual token pruning algorithm, while $heavy \ ratio$ and $recent \ ratio$ are set to $0.5$ and $0.5$ in visual-text interactive token pruning algorithm. It should be noted that our visual pruning method is added after the 16th layer of the CLIP visual encoder. The LoRA parameters are based on the official parameters provided by \textbf{LLaVA1.5-7B}.

\subsection{Ablation Study}
\textbf{Efficiency for Dynamic Visual Token Pruning\quad}
To verify the effectiveness of our proposed dynamic pruning method, we conducted comparative experiments on POPE, TextVQA, and MME datasets with the method of taking the topk similarity. 

\begin{table}[!htb]
\centering
\caption{The effectiveness of the dynamic visual token pruning.}
\label{tab: dynamic threshold}
\begin{tabular}{@{}lllllll@{}}
\toprule
Method & \multicolumn{2}{l}{POPE} & \multicolumn{2}{l}{TextVQA}     & \multicolumn{2}{l}{MME}      \\ \midrule
    Topk  & \multicolumn{2}{l}{86.5 (218)} & \multicolumn{2}{l}{54.97 (272)} & \multicolumn{2}{l}{1417.5 (222)}          \\ \midrule
    Ours    & \multicolumn{2}{l}{\textbf{86.5 (195)}} & \multicolumn{2}{l}{\textbf{55.28 (251)}} & \multicolumn{2}{l}{\textbf{1426.5 (200)}}   \\ \bottomrule
\end{tabular}
\vspace{-5mm}
\end{table}

\begin{table}[!htb]
\centering
\vskip 0.15in
\caption{The effectiveness of visual-text interactive token pruning.}
\label{tab: visual-text}
\begin{tabular}{@{}lllllll@{}}
\toprule
Method & \multicolumn{2}{l}{POPE} & \multicolumn{2}{l}{TextVQA}     & \multicolumn{2}{l}{MME}      \\ \midrule
    Visual  & \multicolumn{2}{l}{86.5 (196)} & \multicolumn{2}{l}{55.28 (254)} & \multicolumn{2}{l}{1426.5 (201)}          \\ \midrule
    Visual + Text   & \multicolumn{2}{l}{\textbf{86.5 (196)}} & \multicolumn{2}{l}{\textbf{55.26 (254)}} & \multicolumn{2}{l}{\textbf{1430.5 (201)}}  \\ \bottomrule
\end{tabular}
\vspace{-5mm}
\end{table}

\begin{table*}
\centering
\caption{Performance comparison with Training-Free and Fine-tuning.}
\label{tab: tvsf}
\begin{tabular}{lcccccc}
\hline
Method         & ScienceQA  & POPE  & MMBench        & VQAv2      & TextVQA    & MME \\ 
  
\hline
Training-Free & 69.75      & \textbf{86.50}    & 61.30    & 75.50      & 55.26  & 1430.50     \\
\hline
Fine-tuning          & \textbf{70.15}      & 85.50    & \textbf{66.00}    & \textbf{76.80}      & \textbf{56.61}   & \textbf{1488.50}   \\ \hline
\end{tabular}
\end{table*}
\vspace{-0.2cm}
The results are shown in Table \ref{tab: dynamic threshold}. The number in parentheses represents the number of visual tokens used, and the original total number of visual tokens is $576$. On the TextVQA and MME datasets, the dynamic pruning method not only outperforms the fixed TopK method in terms of performance but also excels in efficiency. Specifically, on the POPE dataset ours achieves the same performance with fewer tokens compared to the fixed TopK method. These results align with our findings that only a small portion of visual tokens are effective, suggesting redundancy within visual tokens. This also underscores the effectiveness of the proposed dynamic visual token pruning method.

\textbf{Efficiency for Visual-Text Interactive Token Pruning\quad}
Similarly, to demonstrate the effectiveness of our proposed visual-text interactive token pruning method, we conducted comparative experiments with only visual dynamic pruning. The experimental results are shown in Table \ref{tab: visual-text}.
The number in parentheses represents the total number of tokens used (i.e. visual + textual tokens).

\begin{table}[!htb]
\centering
\caption{The effectiveness of visual-text interactive token pruning in long answer dataset.}
\label{tab: newvisual-text}
\resizebox{0.48\textwidth}{!}{
\begin{tabular}{lccccc}
\toprule
Method & \multicolumn{2}{c}{MM-Vet} & \multicolumn{2}{c}{LLaVA-Bench-in-the-Wild}         \\ \midrule
LLaVA  & \multicolumn{2}{c}{30.5 (714)} & \multicolumn{2}{c}{67.8 (772)}           \\ \midrule
    Visual  & \multicolumn{2}{c}{32.2 (266)} & \multicolumn{2}{c}{64.9 (355)}           \\ \midrule
    Visual + Text   & \multicolumn{2}{c}{\textbf{31.5 (191)}} & \multicolumn{2}{c}{\textbf{66.2 (192)}}  \\ \bottomrule
\end{tabular}}
\vspace{-5mm}
\end{table}

\begin{table*}
\centering
\caption{Performance comparison with other multimodal models and pruning methods.}
\label{tab: result}
\begin{tabular}{lcccccc}
\hline
Method         & ScienceQA  & POPE  & MMBench & VQAv2      & TextVQA    & MME \\ \hline
BLIP-2         & 61.00      & 85.30 & -       & 41.00      & 42.50      & 1293.80       \\
InstrucBILP    & 63.10      & 78.90 & -       & -          & 50.70      & 1212.80     \\
Shikra         & -          & -     & 58.80   & 77.40      & -          & -    \\
IDEFICS-9B     & -          & -     & 48.20   & 50.90      & 25.90      & -     \\
IDEFICS-80B    & -          & -     & 54.50   & 60.00      & 30.90      & -     \\
InternVL2-1B   &89.30       &87.30  & 61.60   &70.87       & \textbf{70.50}     &1346.19\\
Qwen-VL        & 67.10      & -     & 38.20   & 78.80      & 63.80       & -    \\
Phi3v          & \textbf{90.80}      &85.80  &\textbf{73.71}    & 75.77          & 69.59      & 1439.12 \\
LLaVA-1.5      & 68.40      &86.40    &66.10    &79.10      & 58.20   &1476.90    \\ 
LLaVA-1.6 &72.80 &\textbf{86.70} &68.13 &\textbf{80.34} &65.70 &\textbf{1498.00} \\
\hline
\textcolor{gray}{Training-Free Method} \\
ToMe &50.00 &52.50 &43.70 &57.10 &45.30 &1138.00\\
LLaVA-PruMerge & 68.52      & 70.70    & 56.78    & 65.90      & 53.51   & 1191.50     \\ 
Recovery Compression &69.01 &72.00 &57.90 &70.41 &55.51 &1284.90 \\ \hline
\textbf{InternVL2-1B + Ours} &88.20 &\textbf{86.87} &61.80 &69.00 &\textbf{63.84} &1347.07 
\\
\textbf{Phi3v + Ours} &\textbf{90.85} &82.78  &\textbf{72.16} & 73.49 &64.79 &1423.86 \\
\textbf{LLaVA-1.5 + Ours}           & 69.75      & 86.50    & 61.30    & 75.50      & 55.26   & 1430.50\\
\textbf{LLaVA-1.6 + Ours} &79.20 &85.50 &67.50 &\textbf{79.30} &56.30 &\textbf{1496.00} \\
 \hline
\textcolor{gray}{Fine-tuning Method} \\
LLaVA-PruMerge+ & 68.30      & 84.00    & 64.90    & 76.80      & \textbf{57.10}   & 1462.40     \\
Recovery Compression &68.72 &79.50 &59.20 &71.18 &56.16 &1323.54\\
CrossGET & 66.70      & 83.90    & 64.70    & \textbf{77.30}      & 54.90   & \textbf{1510.20}    \\ 
\hline
\textbf{LLaVA-1.5-7B + Ours}           & \textbf{70.15}      & \textbf{85.50}    & \textbf{66.00}    & 76.80      & 56.61   & 1488.50
\\ \hline
\end{tabular}
\vspace{-4mm}
\end{table*}

The experimental results indicate that there is minimal performance difference between the two settings. Upon analyzing the datasets, it was found that this is due to all answers in the dataset being single-word answers, which results in a small stride for visual-text pruning. However, in practical, answers in MM-LLMs are predominantly long sentences. To further validate our method, we conducted ablation experiments on two datasets with long answers: MM-Vet \cite{yu2023mm} and LLaVA-Bench-in-the-Wild \cite{liu2024visual}. The result is shown in Table \ref{tab: newvisual-text}, demonstrate that visual-text interactive pruning achieves highly competitive performance while significantly reducing the total number of tokens. Moreover, on the LLaVA-Bench-in-the-Wild dataset, our method surpasses approaches that solely prune visual tokens while using fewer tokens. In conclusion, our proposed method can further reduce computational costs while ensuring performance levels.

\textbf{Fine-tuning VS Training-Free\quad}
Due to the simplicity of our method and its lack of additional parameters, it can be seamlessly integrated into multimodal models at any stage, including both the inference and fine-tuning phases. We compared the performance of the method in two scenarios: the inference stage without training and the fine-tuning. The experimental results are shown in Table \ref{tab: tvsf}. The experimental results show that fine-tuning the model together with our method can further improve the performance. This aligns with common sense and intuition, as the fine-tuning process aids the LLM in learning to understand the pruned tokens, thereby boosting the model's performance even more with fewer tokens.

\subsection{Main Result}
We conducted comparative experiments with other MM-LLMs and existing MM-LLM token pruning methods. The results are shown in Table \ref {tab: result}.
The experimental results show that under the training-freed setting, our proposed method performs best with five base models. As the model has not been fine-tuned to adapt to the pruned tokens, in order to balance performance and efficiency, our average compression rate has reached 22\% $\sim$ 80\% (different datasets and models have varying compression rates).
Under the fine-tuning setting, our method reaches optimal performance on the ScienceQA, POPE, and MMBench datasets, and competitive performance on the VQAv2, TextVQA, and MME datasets with the 7.5 times compression ratio. Compared to LLaVA-PruMerge+, our method offers higher computational efficiency (visual tokens: 127 vs 144), and since we also prune text tokens, we hold an advantage in the total token count. In comparison to CrossGET, our method also demonstrates efficiency advantages. 

Fine-tuning enables the model to adapt to pruned visual tokens, and it may achieve better performance on tasks like ScienceQA and MME datasets. Overall, our proposed method strikes a balance between performance and efficiency, showcasing strong competitiveness and potential. Compared with Recovery Compression, we avoid the situation of excessive compression by using more tokens to ensure that the performance of the model does not significantly decrease in various tasks.

To further evaluate the computational efficiency of our proposed method, we employed open-source simulation software \cite{yuan2024llm} to simulate the computational efficiency on an NVIDIA A100 GPU. During the reverse simulation calculations, we assumed $60$ text tokens as input and the average number of visual tokens used by each method across the six datasets as the input visual token count. The experimental results are shown in Table \ref{tab: costs}. 

\begin{table*}[!htb]
\centering
\caption{Comparison of computational costs on NVIDIA A100 GPU.}
\label{tab: costs}
\begin{tabular}{llccccc}
\hline
Method &
  \multicolumn{1}{c}{\begin{tabular}[c]{@{}c@{}}LLM \\ Backbone\end{tabular}} &
  Quantization &
  \begin{tabular}[c]{@{}c@{}}FLOPs\\ (T)\end{tabular} &
  \begin{tabular}[c]{@{}c@{}}Prefill\\ Time (ms)\end{tabular} &
  \begin{tabular}[c]{@{}c@{}}Total\\ Memory (G)\end{tabular} &
  \begin{tabular}[c]{@{}c@{}}Storing\\ Activation (G)\end{tabular} \\ \hline
LLaVA1.5 & \multicolumn{1}{c}{Vicuna-7B} & FP16 & 8.5          & 30.3         & 22.2          & 4.1           \\
Ours     & \multicolumn{1}{c}{Vicuna-7B} & FP16 & \textbf{2.3} & \textbf{9.7} & \textbf{13.2} & \textbf{0.8} \\ \hline
LLaVA1.5 & Vicuna-7B                     & INT8 & 4.3          & 15.2         & 11.1          & 2.0           \\
Ours     & Vicuna-7B                     & INT8 & \textbf{1.2} & \textbf{4.9} & \textbf{6.6}  & \textbf{0.4} \\ \hline
LLaVA1.5 & Vicuna-7B                     & INT4 & 2.1          & 14.2         & 5.6          & 1.0           \\
Ours     & Vicuna-7B                     & INT4 & \textbf{0.6} & \textbf{2.4} & \textbf{3.3}  & \textbf{0.2} \\ \hline
\end{tabular}%
\end{table*}

Compared to baseline methods, our approach demonstrates significant advantages in computational efficiency. Additionally, when compared to other pruning methods with similar performance levels, our method also exhibits slightly higher computational efficiency. This means that our proposed method can effectively enhance computational efficiency while maintaining performance levels.

\begin{figure}[!htb]
\vspace{-2mm}
\centering
\includegraphics[width=1.0\linewidth]{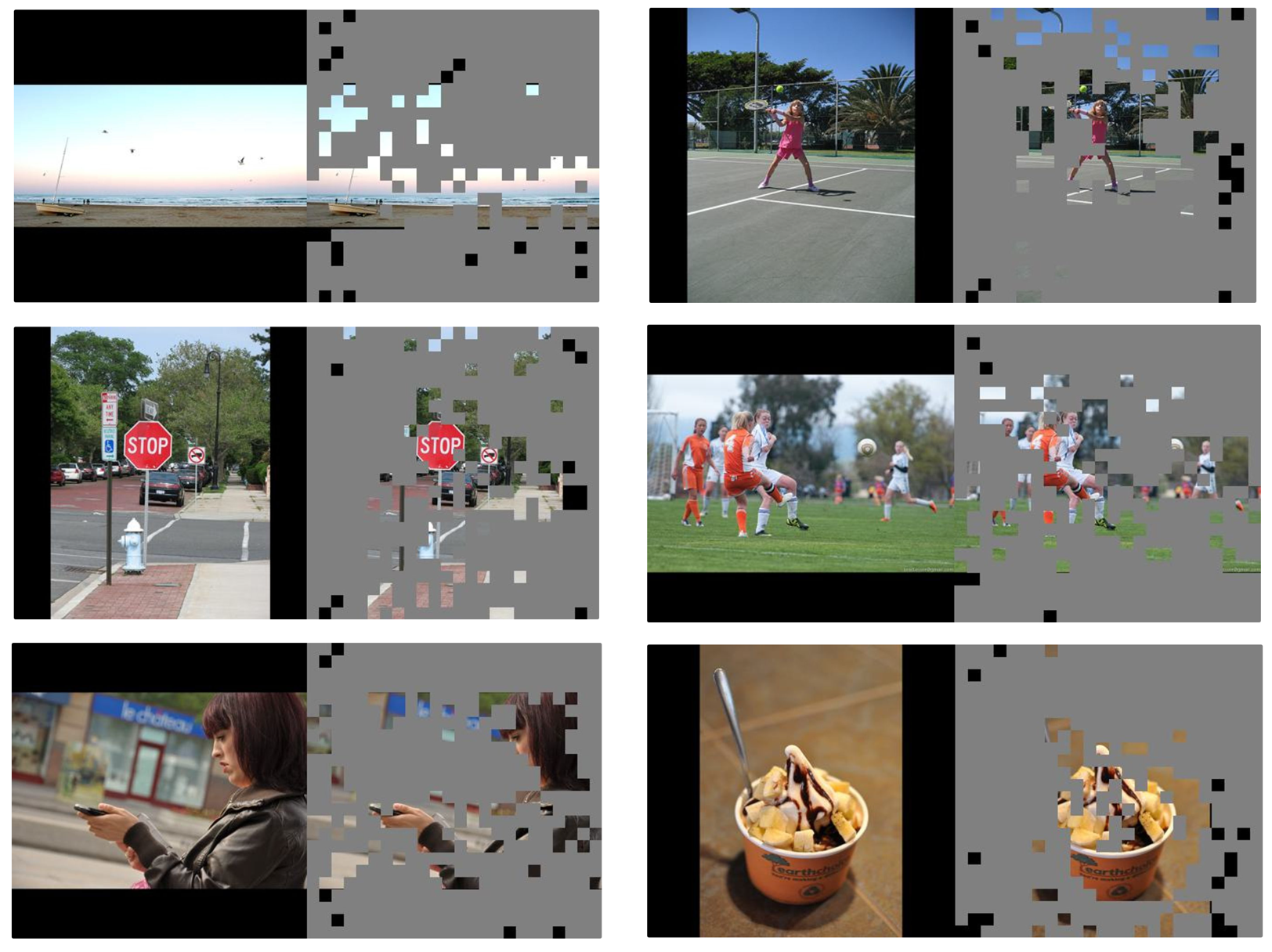}
\caption{The example of Visual Token Visualization Results.} 
\label{fig: v}
\end{figure}

\subsection{Visualization}
To visually illustrate our method, we conducted visualization experiments, as shown in Figure \ref{fig: v}. The grey area represents the regions corresponding to pruned visual tokens, while the areas with normal pixels represent the regions corresponding to retained visual tokens.  

The visualization results demonstrate that visual tokens corresponding to background regions have been mostly discarded, while the retained visual tokens are focused on the main objects in the image (such as boats, people, signs, ice cream, etc.). This indicates that our proposed method can guide the model to concentrate on key areas of the image, effectively enhancing computational efficiency.

\section{Conclusion}
In this paper, we propose a dynamic optimization algorithm for multimodal large language models token compression targeting long-tail distribution. Firstly, we analyze the similarity of CLS visual tokens in the visual encoder, which exhibits a long tail distribution. Therefore, we design an objective function to segment the head and tail in the long tail distribution to extract important visual tokens. Subsequently, the similarity between all tokens is calculated at the LLM layer, and the importance score of each token is obtained by summing up the similarity between the current token and other tokens. Tokens with high importance scores are retained, while tokens with low importance scores are discarded to complete interactive pruning. 

The experimental results on multiple datasets showed that our method achieved the best performance on most datasets and maintained a balance between performance and computational efficiency. This provides ideas and methods for future research on efficient multimodal large models.

\section{Limitation}
In terms of computational efficiency, there is still room for improvement in our method. Further exploration is needed to more accurately capture key visual tokens. Additionally, directly removing background information may potentially harm the model's performance on certain tasks since the background also contains information relevant to problem-solving. In the future, we will continue researching these two areas to enhance the performance of our method further.

\nocite{langley00}

\bibliography{example_paper}
\bibliographystyle{icml2025}

\newpage
\appendix
\onecolumn
\section{Ablation Study}
\subsection{Efficiency for Different Compression Ratios}
We conducted compression ratios ablation experiments on LLaVA-1.5 and InternVL2, and the experimental results are shown in Table \ref{tab: llava} and Table \ref{tab: vl2}.
The experimental results indicate that a compression ratio of around 20\% is a suitable choice, as it can significantly reduce visual tokens while ensuring that performance does not decrease substantially.

\begin{table}[!htb]
\centering
\caption{The impact of different compression ratios on LLaVA-1.5.}
\label{tab: llava}
\begin{tabular}{@{}ccccccc@{}}
\toprule
Compression Ratio & \multicolumn{2}{c}{POPE} & \multicolumn{2}{c}{TextVQA}     & \multicolumn{2}{c}{MME}      \\ \midrule
    10\%   & \multicolumn{2}{c}{78.72} & \multicolumn{2}{c}{54.28} & \multicolumn{2}{c}{1330.63}          \\ 
    20\%   & \multicolumn{2}{c}{83.35} & \multicolumn{2}{c}{\textbf{55.13}} & \multicolumn{2}{c}{1415.71}  \\ 
    50\%  & \multicolumn{2}{c}{\textbf{85.70}} & \multicolumn{2}{c}{54.99} & \multicolumn{2}{c}{\textbf{1486.40}}  \\
    \bottomrule
\end{tabular}
\end{table}

\begin{table}[!htb]
\centering
\caption{The impact of different compression ratios on InternVL2.}
\label{tab: vl2}
\begin{tabular}{@{}ccccccc@{}}
\toprule
Compression Ratio & \multicolumn{2}{c}{POPE} & \multicolumn{2}{c}{TextVQA}     & \multicolumn{2}{c}{MME}      \\ \midrule
    10\%   & \multicolumn{2}{c}{85.86} & \multicolumn{2}{c}{58.05} & \multicolumn{2}{c}{1345.35}          \\ 
    20\%   & \multicolumn{2}{c}{86.69} & \multicolumn{2}{c}{59.50} & \multicolumn{2}{c}{1346.96}  \\ 
    50\%  & \multicolumn{2}{c}{\textbf{86.87}} & \multicolumn{2}{c}{\textbf{63.84}} & \multicolumn{2}{c}{\textbf{1347.07}}  \\
    \bottomrule
\end{tabular}
\end{table}



\end{document}